\documentclass{article}

\usepackage{microtype}
\usepackage{graphicx}
\usepackage{multirow}
\usepackage{array}
\usepackage{xcolor}
\usepackage{caption}
\usepackage{subcaption}
\usepackage{booktabs} 

\usepackage{hyperref}

\newcolumntype{C}[1]{>{\centering\arraybackslash}m{#1}}

\usepackage[preprint]{icml2026}

\usepackage{amsmath}
\usepackage{amssymb}
\usepackage{mathtools}
\usepackage{amsthm}

\usepackage[capitalize,noabbrev]{cleveref}

\theoremstyle{plain}

\theoremstyle{definition}

\theoremstyle{remark}

\usepackage[textsize=tiny]{todonotes}
\usepackage{algorithm}
\newcommand{\STATEV}{\item[]\hspace{1.5em}}
\icmltitlerunning{Application-Driven Pedagogical Knowledge Optimization of Open-Source LLMs via RL and SFT}

\begin{document}

\twocolumn[
  \icmltitle{Application-Driven Pedagogical Knowledge Optimization of Open-Source LLMs \\
    via Reinforcement Learning and Supervised Fine-Tuning}



  \icmlsetsymbol{equal}{*}

  \begin{icmlauthorlist}
    \icmlauthor{Navan Preet Singh$^*$}{compl1}
    \icmlauthor{Xiaokun Wang$^*$}{compl2}
    \icmlauthor{Anurag Garikipati}{compl3}
    \icmlauthor{Madalina Ciobanu}{compl3}
    \icmlauthor{Qingqing Mao}{compl3,compl4,compl1}
    \icmlauthor{Ritankar Das}{compl3,compl4}
  \end{icmlauthorlist}

  \icmlaffiliation{compl1}{Forta, Houston, TX}
  \icmlaffiliation{compl2}{East China Normal University, Shanghai, China}
  \icmlaffiliation{compl3}{Incept Labs, Houston, TX}
  \icmlaffiliation{compl4}{Titan Holdings, San Francisco, CA}

  \icmlcorrespondingauthor{Qingqing Mao}{qmao@inceptlabs.ai}

  \icmlkeywords{Machine Learning, ICML}

  \vskip 0.3in
]
{\renewcommand{\thefootnote}{}\footnotetext{$^*$These authors contributed equally to this work and share first authorship.}}



\printAffiliationsAndNotice{}  

\begin{abstract}
  We present an innovative multi-stage optimization strategy combining reinforcement learning (RL) and supervised fine-tuning (SFT) to enhance the pedagogical knowledge of large language models (LLMs), as illustrated by EduQwen 32B-RL1, EduQwen 32B-SFT, and an optional third-stage model EduQwen 32B-SFT-RL2: (1) RL optimization that implements progressive difficulty training, focuses on challenging examples, and employs extended reasoning rollouts; (2) a subsequent SFT phase that leverages the RL-trained model to synthesize high-quality training data with difficulty-weighted sampling; and (3) an optional second round of RL optimization. EduQwen 32B-RL1, EduQwen 32B-SFT, and EduQwen 32B-SFT-RL2 are an application-driven family of open-source pedagogical LLMs built on a dense Qwen3-32B backbone. These models remarkably achieve high enough accuracy on the Cross-Domain Pedagogical Knowledge (CDPK) Benchmark to establish new state-of-the-art (SOTA) results across the interactive Pedagogy Benchmark Leaderboard and surpass significantly larger proprietary systems such as the previous benchmark leader Gemini‑3 Pro. These dense 32-billion-parameter models demonstrate that domain-specialized optimization can transform mid-sized open-source LLMs into true pedagogical domain experts that outperform much larger general-purpose systems, while preserving the transparency, customizability, and cost-efficiency required for responsible educational AI deployment.
\end{abstract}

\section{Introduction}

Large Language Models (LLMs), driven by deep learning algorithms, are sophisticated systems designed to synthesize knowledge and engage in nuanced natural conversations. These models demonstrate remarkable proficiency in generating coherent, contextually rich output, a capability that has positioned them as transformative instruments across various high-stakes industries (e.g., finance, law, and medicine). While premium proprietary LLMs (e.g., GPT-5, Claude, and Gemini) have historically set general state-of-the-art (SOTA) performance benchmarks \cite{Minaee2024}, advanced open-source LLMs—such as Mistral 3, DeepSeek-R1, and Qwen3—offer significant advantages in research and development due to their publicly accessible architectures and algorithmic transparency \cite{Maharjan2024,Manchanda2025}. A key downside of commercial models is the difficulty and cost associated with fine-tuning, owing to their opacity as well as significantly larger size compared to smaller, specialized open-source models \cite{Liao2023, Wu2024}. 

In the field of pedagogy, LLMs offer the potential for scalable, personalized instruction; while it is currently cost-prohibitive, one-on-one tutoring delivers significantly better educational outcomes (Bloom's Two-Sigma Problem) \cite{Bloom1984}. Open-source LLMs are particularly attractive for implementation in education; when compared with commercial models, they allow for easier fine-tuning for specific requirements (e.g., pedagogy), which enables the creation of more affordable customized tools that can adapt dynamically to individual learner needs \cite{Perczel2025, Yan2024}. The opacity of commercial, proprietary LLMs limits external scrutiny and customization \cite{Liao2023}, presenting a significant barrier to effectively addressing pedagogical steering and bias mitigation. Additionally, commercial models often incur high, usage-based fees and restrict control over data and deployment; the commercial models can also update without notice, leading to development instability. Conversely, the full control and transparency afforded by open-source models facilitate the community-wide scrutiny and domain-specific engineering necessary for transparency, ethical adherence, and robust accountability \cite{NTIA2024}. These considerations are especially critical for educational applications, where deployment at scale directly impacts learning outcomes for diverse student populations.

LLMs can open the door for global personalized education, however, currently these models are typically optimized to prioritize immediate helpfulness \cite{Puech2024}. This leads to a misalignment with the concept of guided learning in which the goal is not to provide the student with the answer, but rather help them get to the answer themselves. This structural gap underscores the importance of evaluating LLMs with specialized frameworks that explicitly distinguish between content knowledge (factual understanding) and pedagogical knowledge (strategic teaching capacity). A recent study introduced The Pedagogy Benchmark—a publicly available dataset of over 1,100 teacher exam questions—demonstrating that commercial and open-source LLMs achieve wide-ranging pedagogical knowledge scores from 21 to 91\%; interactive leaderboards are published in order to transparently compare the performance, cost, and reasoning abilities on pedagogical tasks of LLMs from a growing list of 150 \cite{AIforEducation2026, Lelievre2025}.

To address the urgent need for widely-available and validated educational AI, we strategically enhanced the cross-domain pedagogical knowledge of open-source models, as tested on the Cross-Domain Pedagogical Knowledge (CDPK) Benchmark subset of The Pedagogy Benchmark. Our work leverages the transparency and fine-tuning capacity of open-source models through an innovative multi-stage optimization combining reinforcement learning (RL) and supervised fine-tuning (SFT), thereby advancing the development of trustworthy and accessible AI tutors. In this study, we present a model optimization strategy that establishes new state-of-the-art (SOTA) performance on the CDPK Benchmark. Remarkably, our approach, which applies advanced model training and robust inference-time compute techniques to the dense 32-billion parameter open-source model Qwen3-32B, achieves up to 96.52\% accuracy, surpassing all open-source and proprietary models currently listed on the interactive Pedagogy Benchmark leaderboard \cite{AIforEducation2026}—including significantly larger models—demonstrating that domain-specialized smaller models can outperform general-purpose large-scale systems while offering superior cost-efficiency.

This paper makes three contributions to educational applications of LLMs: (1) the development of EduQwen 32B-RL1, EduQwen 32B-SFT, and the optional third-stage model EduQwen 32B-SFT-RL2 as open-source pedagogical domain experts; (2) a novel multi-stage pipeline that combines pedagogically aligned RL with difficulty-weighted SFT using RL-generated synthetic data; and (3) empirical evidence that this approach achieves new SOTA performance on the CDPK Benchmark, surpassing much larger proprietary models while remaining practical for real-world educational deployment. Our study is therefore positioned primarily as an application-driven contribution in the education domain, demonstrating how targeted optimization can turn open-source LLMs into high-precision pedagogical tutors suitable for classroom and platform integration.

\section{Methodology}

\subsection{Benchmark Evaluation}

The evaluation of LLM performance was conducted exclusively using the CDPK Benchmark, to ensure domain-specific validation of pedagogical knowledge. The CDPK Benchmark is a specialized dataset designed to explicitly distinguish between content knowledge and strategic pedagogical knowledge \cite{Lelievre2025}, which is crucial for informing the responsible deployment of LLMs in educational settings. The CDPK Benchmark comprises a publicly available collection of 920 teacher multiple-choice exam questions, ensuring a rigorous assessment of the models’ capacity to support effective teaching practices.

Model performance was primarily evaluated by calculating accuracy—the percentage of correct responses against a ground truth answer key—a standard metric for static, ground-truth-based educational assessments. The final comparative performance of the tested open-source LLMs was assessed against the interactive Pedagogy Benchmark leaderboard \cite{AIforEducation2026}. This methodology provides a transparent and standardized comparison of the developed platform against a growing list of open-source and commercial LLMs, crucial for benchmarking pedagogical LLM performance.

\subsection{Base Model Selection}

We selected Qwen3-32B as our base model based on systematic evaluation of diverse open-source LLMs to determine the optimal foundation for pedagogical optimization, as illustrated by the performance on the CDPK Benchmark. Qwen3-32B, a dense model developed by Alibaba \cite{Yang2025}, was chosen for its strong baseline performance on the CDPK Benchmark and demonstrated amenability to advanced fine-tuning methods. The Qwen3-32B foundation model is a highly competitive open-source LLM derived from the Qwen3 series architecture, providing the transparency and flexibility that are particularly advantageous for educational deployment while maintaining computational efficiency through its dense 32-billion parameter architecture. We evaluated models spanning both dense and mixture of experts (MoE) architectures in preliminary experiments, finding that dense architectures proved more responsive to iterative optimization \cite{Kim2025, Wang2024}. This rigorous evaluation identified Qwen3-32B as an optimal base LLM for pedagogical specialization, forming the foundation upon which we construct EduQwen 32B-RL1 via RL optimization and EduQwen 32B-SFT via subsequent supervised fine-tuning.

\subsection{Model Training}

This paper proposes a multi-stage iterative training pipeline, RL-SFT-RL, to enhance the reasoning capabilities of LLMs in the pedagogy domain. The process begins with an initial or first round of RL, where a base model (Qwen3-32B) is trained to identify and focus on challenging problems. Subsequently, an SFT phase is introduced to refine the model's proficiency. Finally, an optional second round of RL is conducted to solidify the gains.

\subsubsection{RL OPTIMIZATION}

We introduce an innovative RL optimization strategy that transforms the general-purpose Qwen3-32B into EduQwen 32B-RL1, a domain expert specializing in pedagogical knowledge. Our ablation studies compared Group Relative Policy Optimization (GRPO) \cite{Shao2024} and Decoupled Advantage Policy Optimization (DAPO) \cite{Liu2024}. We found that DAPO's ability to decouple the policy update from the advantage estimation provided more stable gradients when training on the complex, multi-step reasoning chains required for pedagogical scaffolding, ultimately yielding superior performance on the CDPK Benchmark, and was therefore selected as our primary RL algorithm. Specifically, our implementation enhances training stability by employing the DAPO estimator for advantage computation and utilizing asymmetric clipping to stabilize policy updates. This asymmetric approach allows for a more robust exploration of high-reward reasoning paths while strictly maintaining a lower bound to prevent catastrophic policy divergence. We utilized RL with a reward model trained to prioritize pedagogically sound responses that guide learners rather than simply providing answers. Our RL training incorporated several novel strategic innovations to maximize performance on the CDPK Benchmark.

We implemented progressive training by difficulty level, beginning with simpler pedagogical questions and gradually increasing complexity as the model's performance improved. This curriculum learning approach \cite{Bengio2009} allowed the model to build foundational pedagogical reasoning before tackling more nuanced teaching scenarios. Throughout training, we continuously increased the number of epochs, monitoring validation performance to determine optimal stopping points that balanced improvement against overfitting.

Our data processing strategy explicitly targeted the model's weaknesses through a rigorous failure-mode identification process. We first tested the Qwen3-32B base model on all questions in the pedagogy dataset, performing 30 attempts per question to identify items where the model exhibited uncertainty or incorrect reasoning. Questions answered correctly in all 30 attempts were excluded from training. The remaining questions were sorted by error frequency from lowest to highest, creating a difficulty-ordered curriculum of 440 training data points that progressed from easier to more challenging pedagogical scenarios. This carefully curated dataset formed the foundation for our first-stage RL optimization, ensuring that training focused precisely on the model's areas of greatest need. This selection method effectively serves as an importance-sampling mechanism, where the synthetic data distribution is shifted toward the hard-negative space of the RL-optimized policy. By training on these high-difficulty samples, the SFT phase acts as a corrective regularizer that smooths the policy's performance across the entire pedagogical domain.

A critical innovation in our approach was training on "potentially incorrect questions"—examples where the model's initial responses were uncertain or showed reasoning flaws. By generating synthetic training data focused on these failure modes, similar to techniques in the Self-Taught Reasoner (STaR) \cite{Zelikman2022}, we enabled the model to strengthen its weakest areas iteratively. We also progressively increased rollout length from 5 to 8 steps during RL training, allowing the model to engage in longer chains of reasoning and better capture the multi-step nature of pedagogical decision-making. This innovative alignment process resulted in our RL-optimized model, EduQwen 32B-RL1, achieving 94.13\% accuracy on the CDPK Benchmark, already establishing new SOTA performance for both open-source and proprietary models in educational pedagogy.

Following the initial RL stage and subsequent SFT phase (described in Section 2.3.2), we applied an optional second round of RL optimization to further refine pedagogical reasoning capabilities. This second round of RL training utilized the same difficulty-ordered data points from the first RL stage, maintaining the progressive difficulty arrangement to reinforce the model's performance on challenging pedagogical items while building upon the improvements achieved through SFT.

\subsubsection{SFT}

Following first-stage RL optimization, we apply a novel SFT strategy that converts EduQwen 32B-RL1 into an intermediate model (EduQwen 32B-SFT) during a second stage, further enhancing pedagogical performance and establishing new SOTA. The intermediate model can optionally undergo a third-stage optimization with another round of RL (described in subsection 2.3.3.) to produce EduQwen 32B-SFT-RL2, establishing definitive SOTA results for educational LLM applications. The SFT phase leverages the RL-trained model itself (EduQwen 32B-RL1) to synthesize high-quality training data, creating a virtuous cycle of improvement. We prompted the RL-optimized model to generate 40,000 data points consisting of pedagogically sound responses to questions across the difficulty spectrum, then applied a rigorous filtering process to extract high-quality examples focused on the model's challenging areas.

A key innovation in our SFT approach was the strategic weighting of training examples by difficulty level, explicitly targeting high-difficulty pedagogical items that typical educational LLMs struggle with. Our data curation process applied gradient-based selection: we eliminated questions the model answered correctly consistently, retaining only items where the model exhibited uncertainty or errors. For questions with high baseline accuracy, we selected a single representative training example; for genuinely difficult questions with low accuracy, we retained all available high-quality examples to ensure comprehensive coverage of challenging pedagogical scenarios. This process yielded 1,050 high-quality, difficulty-ordered training data points that concentrated learning capacity on the model's areas of greatest need. We assigned increased training weights to high-difficulty questions, ensuring the model developed robust performance on the most challenging scenarios. This difficulty-aware training strategy helped address the typical performance degradation that models experience on complex edge cases.

The SFT process utilized standard cross-entropy loss with the difficulty-weighted examples, training for multiple epochs while monitoring validation accuracy. We employed gradient accumulation and mixed-precision training to optimize computational efficiency. The combination of RL-generated synthetic data and difficulty-weighted training enabled the model to achieve a significant performance gain over the RL-only baseline. The outcome of this stage is EduQwen 32B-SFT, a pedagogical model that establishes new SOTA, and whose performance can be further improved by applying an optional second round of RL to establish definitive SOTA.

\subsubsection{SECOND ROUND RL OPTIMIZATION}

Following the first round of RL optimization and subsequent SFT phase, we optionally apply a second round of RL optimization starting from EduQwen 32B-SFT and using the same difficulty-aware dataset constructed during the first RL round. Crucially, this final RL phase reuses the initial difficulty-aware dataset, allowing the model to leverage its refined knowledge to finally solve the problems it initially found challenging. This cyclical process of reinforcement, refinement, and re-reinforcement ensures a systematic improvement in the model's performance on complex pedagogical tasks, yielding our final model EduQwen 32B-SFT-RL2.

The complete multi-stage optimization procedure—comprising the initial RL stage for hard-negative identification, the difficulty-weighted SFT phase for knowledge distillation, and the optional final RL refinement—is formalized in Algorithm 1. This framework illustrates the iterative transition toward domain expertise, specifically detailing how we stabilize policy updates across all phases by employing the DAPO estimator and expansive asymmetric clipping.

\begin{algorithm}[ht]
\caption{Pedagogy Domain Training via RL-SFT-RL}
\label{alg:rlsft}
\begin{algorithmic}[1]
\REQUIRE Base Model $\theta_0$, Dataset $\mathcal{D}$, Epochs $E_{RL}$, $E_{SFT}$
\ENSURE Fine-tuned Model $\theta_{final}$

\STATEV \textbf{Phase 1: Initial RL (DAPO)}
\STATE $\mathcal{D}_{hard}^{(1)} \gets \text{DifficultyFilter}(\mathcal{D}, \theta_0, \text{threshold}=0)$ \COMMENT{Filter out perfect samples}
\STATE $\theta_{RL1} \gets \text{DAPO\_Train}(\theta_0, \mathcal{D}_{hard}^{(1)}, E_{RL})$ \COMMENT{Using DAPO and asymmetric clipping}

\STATEV \textbf{Phase 2: Synthetic SFT}
\STATE $\mathcal{D}_{syn} \gets \theta_{RL1}.\text{Generate}(N=40000)$ \COMMENT{Generate synthetic data}
\STATE $\mathcal{D}_{clean} \gets \{d \in \mathcal{D}_{syn} \mid \text{Acc}(d) > 0 \}$ \COMMENT{Retain only correct responses}
\STATE $\mathcal{D}_{final}^{SFT} \gets \text{GradientSelection}(\mathcal{D}_{clean})$ \COMMENT{Gradient selection: retain hard}
\STATE $\theta_{SFT} \gets \text{SFT\_Train}(\theta_{RL1}, \mathcal{D}_{final}^{SFT}, E_{SFT})$

\STATEV \textbf{Phase 3: Final RL (DAPO)}
\STATE $\theta_{final} \gets \text{DAPO\_Train}(\theta_{SFT}, \mathcal{D}_{hard}^{(1)}, E_{RL})$ \COMMENT{Re-use Phase 1 hard dataset}

\STATE \textbf{return} $\theta_{final}$
\end{algorithmic}
\end{algorithm}

\section{Results}

Our innovative multi-stage optimization strategy—RL to produce EduQwen 32B-RL1 followed by strategic SFT to obtain EduQwen 32B-SFT, and optionally a second RL round that yields EduQwen 32B-SFT-RL2—results in unprecedented performance on the CDPK Benchmark, establishing new SOTA results. EduQwen 32B-RL1 reaches 94.13\% accuracy, EduQwen 32B-SFT reaches 96.20\% accuracy, and with the optional second RL round, EduQwen 32B-SFT-RL2 reaches 96.52\% accuracy, with all three models surpassing all open-source and proprietary systems currently listed on The Pedagogy Benchmark Leaderboard, including the previous best-performing Gemini-3 Pro at 90.55\%. Table 1 showcases the performance of baseline models and our optimized variants.

\begin{table}[ht]
\caption{Performance of EduQwen variants (EduQwen 32B-RL1, EduQwen 32B-SFT, EduQwen 32B-SFT-RL2) on the CDPK Benchmark compared with baseline and leaderboard models.}
\label{tab:performance}
\vskip 0.15in
\begin{center}
\begin{small}
\begin{sc}
\setlength{\tabcolsep}{2pt} 
\begin{tabular}{>{\raggedright\arraybackslash}p{3.6cm}>{\raggedright\arraybackslash}p{2.15cm}>{\centering\arraybackslash}p{1cm}>{\centering\arraybackslash}p{1cm}}
\toprule
Model & Inference-Time Compute & Repl. Acc. & Rep. Acc. \\
\midrule
Gemini-3Pro (CDPK Rank \#1) & Benchmark Leader & -- & {\color{red}90.55} \\
\addlinespace[3pt]
Qwen3-235B-A22B-Thinking (CDPK Rank \#19) & Random Few-Shot & 86.63 & 85.65 \\
\addlinespace[3pt]
Qwen2.5-72B-Instruct (CDPK Rank \#66) & Random Few-Shot & 78.60 & 75.42 \\
\addlinespace[3pt]
Qwen3-32B Base Model (CDPK Rank \#28) & Random Few-Shot & 83.37 & 82.42 \\
\midrule
\textbf{EduQwen 32B-RL1} & RL 1st Round & \textbf{94.13} & -- \\
\addlinespace[3pt]
\textbf{EduQwen 32B-SFT} & SFT & \textbf{96.20} & -- \\
\addlinespace[3pt]
\textbf{EduQwen 32B-SFT-RL2} & RL 2nd Round & \textbf{96.52} & -- \\
\bottomrule
\end{tabular}
\end{sc}
\end{small}
\end{center}
\vskip -0.1in
\begin{flushleft}
\fontsize{8pt}{9pt}\selectfont \textit{Note:} \textbf{Bold} font indicates the best performance between our EduQwen models and the Cross-Domain Pedagogical Knowledge (CDPK) leader. RL = reinforcement learning; SFT = supervised fine-tuning. Repl. = Replicated. Rep. = Reported. Acc. = Accuracy. Reported accuracy refers to the performance published on the CDPK Benchmark Leaderboard, while replicated accuracy indicates our experimental results.
\end{flushleft}
\end{table}
The performance of EduQwen 32B-RL1, EduQwen 32B-SFT, and EduQwen 32B-SFT-RL2 in comparison to other top-performing models on the CDPK Benchmark Leaderboard \cite{AIforEducation2026} demonstrates that our approach establishes a new SOTA for pedagogical knowledge in LLMs. The substantial improvement from the Qwen3-32B baseline (83.37\%) to our final model (96.52\%) represents a 13.15 point gain, showcasing the effectiveness of our innovative combined RL and SFT optimization pipeline.

Figure 1 illustrates the training dynamics of the RL optimization phase. The reward signal exhibits high variance during early training as the model explores the pedagogical reasoning space, then shows rapid improvement with reward climbing from approximately 0.5 to near 1.0 within the first 100 steps. Convergence is achieved by approximately step 400, with variance decreasing substantially as the model stabilizes on pedagogically-aligned response patterns.

\begin{figure}[ht]
\vskip 0.2in
\begin{center}
\centerline{\includegraphics[width=\columnwidth]{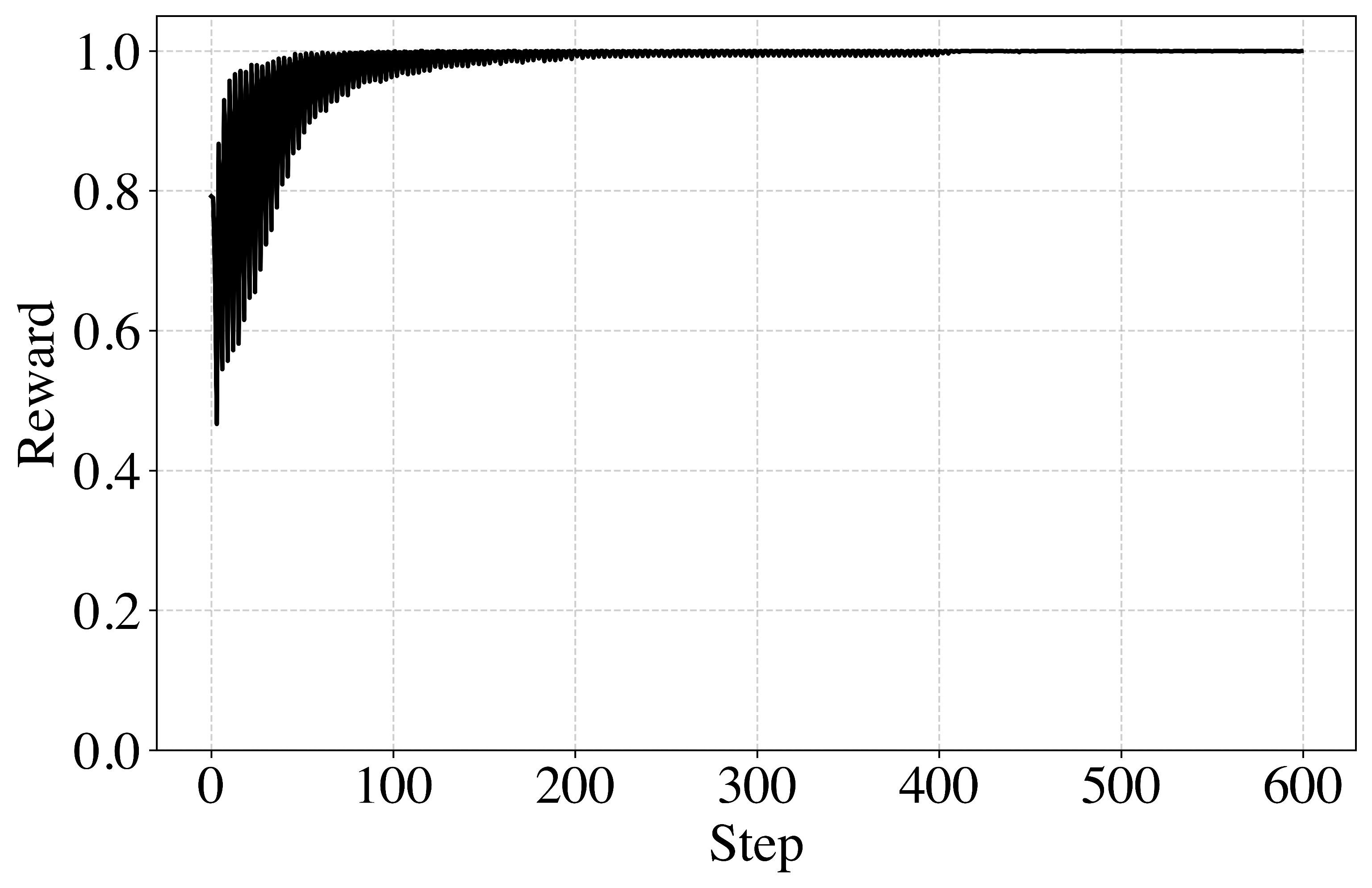}}
\caption{Reward curve during first-stage reinforcement learning optimization of EduQwen 32B-RL1, illustrating convergence across training iterations. }
\label{fig:rl-reward}
\end{center}
\vskip -0.2in
\end{figure}

Figure 2 shows the corresponding SFT training loss curve, which decreases from approximately 0.5 to near zero, with convergence occurring by approximately step 150. The rapid convergence of both training phases suggests that the Qwen3-32B base model is highly amenable to pedagogical specialization, and that our difficulty-weighted curriculum provides an efficient learning signal.

\begin{figure}[ht]
\vskip 0.2in
\begin{center}
\centerline{\includegraphics[width=\columnwidth]{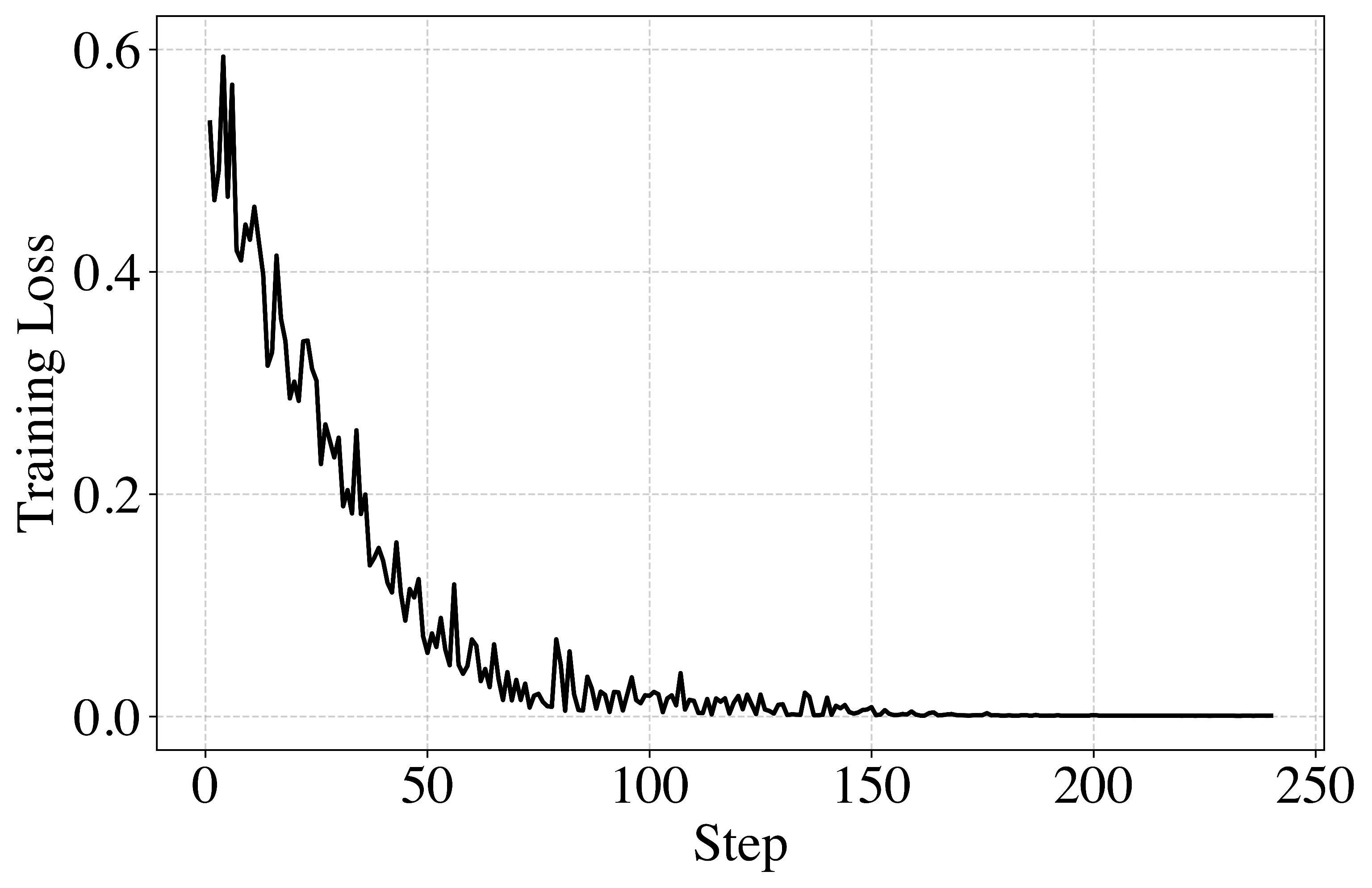}}
\caption{Training loss curve during supervised fine-tuning of EduQwen 32B-SFT using synthetic data generated by the RL-optimized model.}
\label{fig:sft-loss}
\end{center}
\vskip -0.2in
\end{figure}

Notably, all stages of our optimization contributed meaningfully to the final performance: the first RL stage provided a 10.76 percentage point improvement over baseline (83.37\% to 94.13\%), the subsequent SFT phase added an additional 2.07 points (94.13\% to 96.20\%), and the optional second RL round contributed a further 0.32 points (96.20\% to 96.52\%). Experimental records confirm that both RL and SFT processes reached complete convergence (Figures 1 and 2).

This demonstrates that our multi-stage approach captures complementary aspects of pedagogical reasoning—RL for strategic alignment and SFT for knowledge refinement. Critically, our optimization strategy applied to the dense 32-billion parameter model surpasses the performance of significantly larger models such as Qwen3-235B-A22B-Thinking (86.63\%) and Qwen2.5-72B-Instruct (78.60\%), demonstrating that domain-specialized optimization transforms smaller open-source models into true domain experts that are more cost-efficient than running very large models with broad general knowledge.

\section{Discussion}

Our results demonstrate that the EduQwen 32B-RL1, EduQwen 32B-SFT, and optional EduQwen 32B-SFT-RL2 pipeline achieves substantial improvements in LLM performance on specialized pedagogical tasks in education, establishing new SOTA results. The 96.52\% accuracy achieved by our final model (EduQwen 32B-SFT-RL2) represents the highest performance among both open-source and proprietary models currently benchmarked, surpassing Gemini-3 Pro by 5.97 percentage points. These improvements are critical for the development of widely accessible, transparent, and high-performing models for personalized instruction.

The effectiveness of our novel multi-stage optimization pipeline highlights several important insights for domain-specific LLM development in educational applications. First, RL training with pedagogically-aligned reward models successfully shifts model behavior toward guided learning rather than direct answer provision. Second, training on "potentially incorrect questions" and using progressive difficulty curricula enables targeted improvement on model weaknesses. Third, leveraging the RL-optimized model to generate synthetic training data for SFT creates a virtuous cycle that pushes performance beyond what either technique achieves independently.

The combination of strategic innovations during RL training—including rollout step adjustment (e.g., increased rollout length from 5 to 8 steps), the selection of DAPO over GRPO, progressive difficulty curricula, and continuously increasing training epochs—proved collectively valuable for pedagogical reasoning performance, which often requires multi-step consideration of learner needs, appropriate scaffolding, and strategic information revelation. This finding suggests that pedagogical tasks benefit from holistic optimization strategies that enable longer reasoning chains and targeted learning.

A crucial contribution of our work is demonstrating that a dense 32-billion parameter open-source model (Qwen3-32B), when optimized as a domain expert through our innovative training approach, surpasses the performance of significantly larger models including those exceeding 200 billion parameters. This finding has direct implications for educational applications of machine learning: domain-specialized, mid-sized open-source models such as EduQwen 32B-RL1, EduQwen 32B-SFT, and EduQwen 32B-SFT-RL2 can deliver expert-level pedagogical performance at a fraction of the cost of very large proprietary systems. The open-source architecture of our model enables educators and institutions to audit, customize, and adapt the system to specific pedagogical frameworks and cultural contexts—capabilities unavailable with proprietary systems. These characteristics make our approach particularly suitable for real-world educational deployment where institutional constraints often preclude reliance on proprietary APIs.

To assess the generalizability of our optimization approach beyond the CDPK Benchmark, we conducted preliminary experiments on TutorBench \cite{Srinivasa2025}, a recently introduced multimodal benchmark that evaluates LLM tutoring capabilities across both text-only and multimodal educational scenarios. Using Qwen3-30B-VL as a vision-language base model—selected for its vision-language capabilities required by TutorBench's multimodal tasks—we applied a single SFT optimization run using 596 high-quality synthetic responses filtered from model outputs. As shown in Table 2, our SFT-optimized model (EduQwen 30B-VL-SFT) achieves 61.64\% overall accuracy, surpassing Gemini-3 Pro (58.52\%), a top-performing model on the TutorBench leaderboard (Rank no. 3) \cite{Scale2025}, despite starting from a weaker baseline (55.39\%). These preliminary results suggest that our difficulty-weighted SFT methodology transfers effectively across educational benchmarks and modalities, though we note that our evaluation used Claude-4.5-Sonnet as judge compared to Claude-4-Sonnet used in the official leaderboard.


\begin{table}[ht]
\caption{Performance of EduQwen 30B-VL-SFT on the TutorBench Benchmark, compared with baseline and leaderboard models.}
\label{tab:tutorbench-performance}
\vskip 0.15in
\begin{center}
\begin{small}
\begin{sc}
\setlength{\tabcolsep}{2pt} 
\begin{tabular}{>{\raggedright\arraybackslash}p{2.5cm}p{1.3cm}>{\centering\arraybackslash}p{0.9cm}>{\centering\arraybackslash}p{0.9cm}>{\centering\arraybackslash}p{0.9cm}>{\centering\arraybackslash}p{0.9cm}}
\toprule
Model & Judge Model & Text-only & Multi-modal & Over-all & Tutor Bench \\
\midrule
Gemini-2.5-pro-preview-06-05 (Rank \#1) & Claude-4-Sonnet & -- & -- & -- & {\color{red}55.65} \\
\addlinespace[3pt]
Gemini-3-pro-preview (Rank \#3) & Claude-4-Sonnet & -- & -- & -- & 53.67 \\
\addlinespace[3pt]
Gemini-3Pro & Claude-4.5-Sonnet & 58.41 & 58.67 & 58.52 & -- \\
\addlinespace[3pt]
Qwen3-30B-VL (Base Model) & Claude-4.5-Sonnet & 60.21 & 51.53 & 55.39 & -- \\
\midrule
\textbf{EduQwen 30B-VL-SFT} & Claude-4.5-Sonnet & \textbf{64.10} & \textbf{59.40} & \textbf{61.64} & -- \\
\bottomrule
\end{tabular}
\end{sc}
\end{small}
\end{center}
\vskip -0.1in
\begin{flushleft}
\fontsize{8pt}{9pt}\selectfont \textit{Note:} \textbf{Bold} font indicates the best performance between our EduQwen model and the TutorBench top performers. SFT = supervised fine-tuning.
\end{flushleft}
\end{table}
While our current approach achieves new SOTA performance, several opportunities exist for future work. Due to time constraints and associated compute costs, we were unable to test at this time our optimization pipeline on larger open-source models such as DeepSeek-R1 or Qwen3-235B-A22B-Thinking. We hypothesize that our RL and SFT strategies would yield similar improvements when applied to these larger architectures, potentially pushing performance even higher. Additionally, exploring alternative RL algorithms and optimization approaches may offer efficiency gains or performance improvements.

A key limitation of our study is that all primary results are obtained on teacher-exam-style multiple-choice questions from CDPK, so future work should examine free-form tutoring dialogs and longitudinal learning gains to fully validate pedagogical impact. In particular, evaluating EduQwen models in interactive classroom or platform settings would help determine whether benchmark gains translate into sustained improvements in real learners’ reasoning and outcomes.

Our work demonstrates that open-source models can not only match but exceed proprietary models on specialized tasks when equipped with appropriate optimization strategies. This finding has important implications for educational AI deployment, where transparency, customizability, and cost-effectiveness are paramount. The ability to fine-tune open-source models for specific pedagogical frameworks while maintaining full control over model behavior and data addresses key concerns about deploying AI in educational settings.

\section{Conclusion}

In this study, we demonstrate that a novel multi-stage optimization approach combining RL and SFT yields EduQwen 32B-RL1, EduQwen 32B-SFT, and EduQwen 32B-SFT-RL2, which are open-source pedagogical tutors that achieve unprecedented performance on educational tasks, establishing the new SOTA. Our final model achieves 96.52\% accuracy on the CDPK Benchmark, the highest performance among all open-source and proprietary models currently listed on The Pedagogy Benchmark Leaderboard. The combination of RL-based alignment with pedagogical reasoning, progressive difficulty training, focus on challenging examples, and difficulty-weighted SFT created a powerful optimization pipeline that transforms a dense 32-billion parameter model into a true domain expert in pedagogical knowledge. Our results demonstrate that domain-specialized smaller open-source models can surpass significantly larger general-purpose systems, offering superior cost-efficiency while maintaining the critical advantages of transparency, customizability, and adaptability essential for responsible deployment in diverse educational contexts.

\section*{Impact Statement}

This paper presents work aimed at advancing machine learning for educational applications. Our development of high-performing, open-source pedagogical AI systems demonstrates that transparent, customizable AI tutors can achieve expert-level pedagogical knowledge, potentially democratizing access to high-quality personalized education. By achieving SOTA performance with open-source models, we reduce barriers to entry for educational institutions and researchers who lack resources to license expensive proprietary systems. However, deployment of AI tutoring systems raises important concerns about over-reliance potentially reducing human interaction in learning, and careful validation in diverse educational contexts is essential before deployment. We emphasize that our models should augment rather than replace human educators.


\bibliography{example_paper}

@misc{AIforEducation2026,
  author = {{AI for Education}},
  title = {The Pedagogy Benchmark},
  year = {2026},
  url = {https://benchmarks.ai-for-education.org/},
  note = {Retrieved 01/21/2026}
}

@inproceedings{Bengio2009,
  author = {Bengio, Yoshua and Louradour, J{\'e}r{\^o}me and Collobert, Ronan and Weston, Jason},
  title = {Curriculum learning},
  booktitle = {Proceedings of the 26th Annual International Conference on Machine Learning},
  pages = {41--48},
  year = {2009}
}

@article{Bloom1984,
  author = {Bloom, Benjamin S.},
  title = {The 2 sigma problem: The search for methods of group instruction as effective as one-to-one tutoring},
  journal = {Educational Researcher},
  volume = {13},
  year = {1984}
}

@article{Kim2025,
  author = {Kim, J. and Song, M. and Shin, S. and Son, S.},
  title = {Defending {MoE} {LLM}s against Harmful Fine-Tuning via Safety Routing Alignment},
  journal = {arXiv preprint arXiv:2509.22745},
  year = {2025}
}

@article{Lelievre2025,
  author = {Maxime Lelièvre and Amy Waldock and Meng Liu and Natalia Valdés Aspillaga and Alasdair Mackintosh and María José Ogando Portela and Jared Lee and Paul Atherton and Robin A. A. Ince and Oliver G. B. Garrod},
  title = {Benchmarking the Pedagogical Knowledge of Large Language Models},
  journal = {arXiv preprint arXiv:2506.18710},
  year = {2025}
}

@article{Liao2023,
  author = {Liao, Q. Vera and Vaughan, Jennifer Wortman},
  title = {Ai transparency in the age of llms: A human-centered research roadmap},
  journal = {arXiv preprint arXiv:2306.01941},
  year = {2023}
}

@article{Liu2024,
  author = {Jiacai Liu and Chaojie Wang and Chris Yuhao Liu and Liang Zeng and Rui Yan and Yiwen Sun and Yang Liu and Yahui Zhou
},
  title = {Improving Multi-Step Reasoning Abilities of Large Language Models with Direct Advantage Policy Optimization},
  journal = {arXiv preprint arXiv:2412.18279},
  year = {2024}
}

@article{Maharjan2024,
  author = {Jenish Maharjan and Anurag Garikipati and Navan Preet Singh and Leo Cyrus and Mayank Sharma and Madalina Ciobanu and Gina Barnes and Rahul Thapa and Qingqing Mao and Ritankar Das },
  title = {{OpenMedLM}: prompt engineering can out-perform fine-tuning in medical question-answering},
  journal = {Scientific Reports},
  year = {2024}
}

@article{Manchanda2025,
  author = {Manchanda, J. and Boettcher, L. and Westphalen, M. and Jasser, J.},
  title = {The open source advantage in large language models (llms)},
  journal = {arXiv preprint arXiv:2412.12004},
  year = {2025}
}

@article{Minaee2024,
  author = {Shervin Minaee and Tomas Mikolov and Narjes Nikzad and Meysam Chenaghlu and Richard Socher and Xavier Amatriain and Jianfeng Gao},
  title = {Large language models: A survey},
  journal = {arXiv preprint arXiv:2402.06196},
  year = {2024}
}

@misc{NTIA2024,
  author = {{NTIA}},
  title = {Dual-Use Foundation Models with Widely Available Model Weights Report},
  year = {2024},
  url = {https://www.ntia.gov/programs-and-initiatives/artificial-intelligence/open-model-weights-report}
}

@misc{Perczel2025,
  author = {Perczel, J. and Chow, J. and Demszky, D.},
  title = {{TeachLM}: Post-Training {LLM}s for Education Using Authentic Learning Data},
  year = {2025}
}

@article{Puech2024,
  author = {Romain Puech and Jakub Macina and Julia Chatain and Mrinmaya Sachan and Manu Kapur},
  title = {Towards the pedagogical steering of large language models for tutoring},
  journal = {arXiv preprint arXiv:2410.03781},
  year = {2024}
}

@misc{Scale2025,
  author = {{scale.com}},
  title = {{TutorBench}},
  year = {2025},
  url = {https://scale.com/leaderboard/tutorbench}
}

@article{Shao2024,
  author = {Zhihong Shao and Peiyi Wang and Qihao Zhu and Runxin Xu and Junxiao Song and Xiao Bi and Haowei Zhang and Mingchuan Zhang and Y.K. Li and Y. Wu and Daya Guo},
  title = {{DeepSeekMath}: Pushing the Limits of Mathematical Reasoning},
  journal = {arXiv preprint arXiv:2402.03300},
  year = {2024}
}

@article{Srinivasa2025,
  author = {Rakshith S Srinivasa and Zora Che and Chen Bo Calvin Zhang and Diego Mares and Ernesto Hernandez and Jayeon Park and Dean Lee and Guillermo Mangialardi and Charmaine Ng and Ed-Yeremai Hernandez Cardona and Anisha Gunjal and Yunzhong He and Bing Liu and Chen Xing},
  title = {{TutorBench}: A Benchmark To Assess Tutoring Capabilities Of Large Language Models},
  journal = {arXiv preprint arXiv:2510.02663},
  year = {2025}
}

@article{Wang2024,
  author = {Wang, Z. and Chen, D. and Dai, D. and Xu, R. and Li, Z. and Wu, Y.},
  title = {Let the expert stick to his last: Expert-specialized fine-tuning for sparse architectural large language models},
  journal = {arXiv preprint arXiv:2407.01906},
  year = {2024}
}

@article{Wu2024,
  author = {Wu, E. and Wu, K. and Zou, J.},
  title = {{FineTuneBench}: How well do commercial fine-tuning APIs infuse knowledge into {LLM}s?},
  journal = {arXiv preprint arXiv:2411.05059},
  year = {2024}
}

@article{Yan2024,
  author = {Lixiang Yan and Lele Sha and Linxuan Zhao and Yuheng Li and Roberto Martinez-Maldonado and Guanliang Chen and Xinyu Li and Yueqiao Jin and Dragan Gašević},
  title = {Practical and ethical challenges of large language models in education: A systematic scoping review},
  journal = {British Journal of Educational Technology},
  volume = {55},
  number = {1},
  pages = {90--112},
  year = {2024}
}

@article{Yang2025,
  author = {An Yang and Anfeng Li and Baosong Yang and Beichen Zhang and Binyuan Hui and Bo Zheng and Bowen Yu and Chang Gao and  Chengen Huang and Chenxu Lv and Chujie Zheng and Dayiheng Liu and Fan Zhou and Fei Huang and Feng Hu and Hao Ge and Haoran Wei and Huan Lin and Jialong Tang and {...} and Zihan Qiu},
  title = {{Qwen3} technical report},
  journal = {arXiv preprint arXiv:2505.09388},
  year = {2025}
}

@article{Zelikman2022,
  author = {Zelikman, Eric and Wu, Yuhuai and Mu, Jesse and Goodman, Noah D.},
  title = {{STaR}: Bootstrapping Reasoning With Reasoning},
  journal = {arXiv preprint arXiv:2203.14465},
  year = {2022}
}
\bibliographystyle{icml2026}



\end{document}